\documentclass{article}

\usepackage[preprint]{neurips_2026}

\usepackage[utf8]{inputenc} 
\usepackage[T1]{fontenc}    
\usepackage{hyperref}       
\usepackage{url}            
\usepackage{booktabs}       
\usepackage{amsfonts}       
\usepackage{nicefrac}       
\usepackage{microtype}      
\usepackage{xcolor}         

\usepackage{graphicx}
\usepackage{subcaption}
\usepackage{amsmath}
 
\title{Algospeak, Hiding in the Open: The Trade-off Between Legible Meaning and  Detection Avoidance}

\author{Jan Fillies \\
  Stanford University\\
  Stanford, California, USA\\
  Freie Universität Berlin, Germany \\
  \texttt{fillies@stanford.edu} \\\And
  Ronald E. Robertson \\
   Stanford University\\
   Stanford, California, USA\\\And
  Jeffrey Hancock\\
  Stanford University\\
  Stanford, California, USA\\}

\begin{document}

\maketitle

\begin{abstract}
As large language models (LLMs) increasingly mediate both content generation and moderation, linguistic evasion strategies known as Algospeak have intensified the coevolution between evaders and detectors. This research formalizes the underlying dynamics grounded in a joint action model: when Algospeak increases, detectability and understandability decrease. Further, the concept of Majority Understandable Modulation (MUM) is introduced and defined as the modulation level at which additional evasive alteration increases detector evasion but loses comprehension for the majority of recipients. To empirically probe this trade-off, we introduce a reproducible framework that can be used to create meaning-preserving, Algospeak-style variants, based on an existing taxonomy and with tunable modulation levels. Using COVID-19 disinformation as a first proof-by-example setting, we construct a reference dataset of 700 modulated items, drawn from twenty base sentences across five modulation levels and seven strategies. We then run two linked evaluations with seven different language models: one testing for interpretation through meaning recovery and one for disinformation detection through classification. Curve fitting over modulation levels yields an estimate of the Majority Understandable Modulation threshold and enables sensitivity analyses across strategies and models, see Figure \ref{fig:llm_compare}. Results reveal the characteristic relationships between understandability and modulation. This study lays the groundwork for understanding the dynamics behind Algospeak and provides the framework, dataset, and experimental setups described. 
\end{abstract}

\begin{figure}[h]
    \centering
    \includegraphics[width=3in]{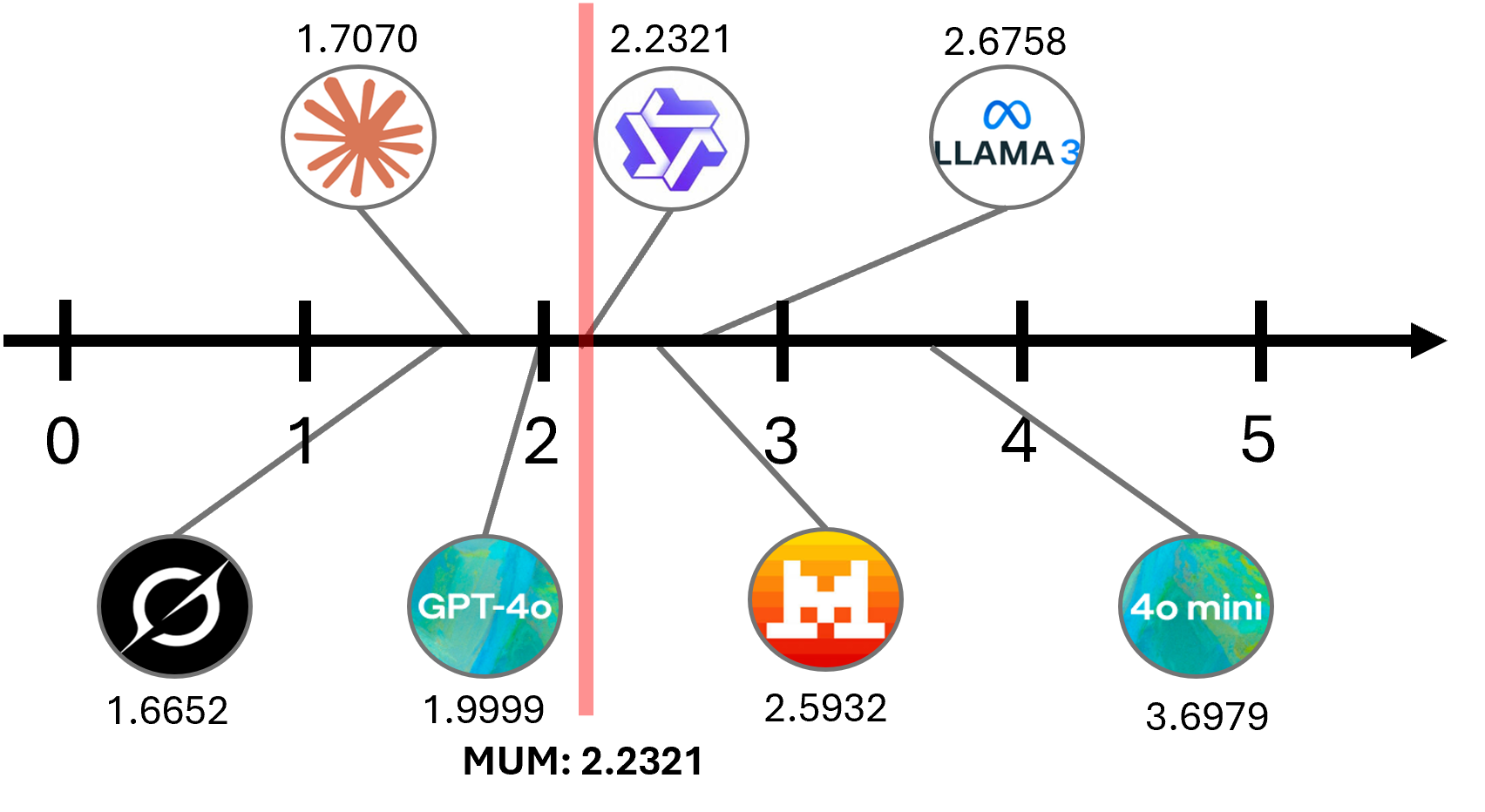}
        \caption{Strategy ``Code Word'': Shows the number of code words required for each model to fail to detect the majority of disinformation (IMUM points). The MUM point is where the majority of models stop detecting accurately.}
    \label{fig:llm_compare}
\end{figure}
\begin{figure*}
    \centering
    \includegraphics[width=4in]{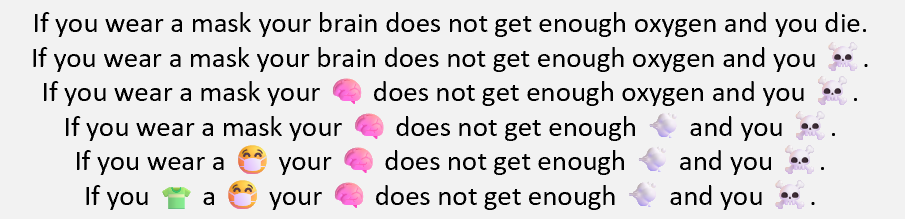}
    \caption{Example misinformation item with five levels of modulation.}
    \label{fig:modulated_ex}
\end{figure*}

\section{Introduction}
As the capabilities of large language models (LLMs) increase, they are gaining traction with malicious actors online. LLMs have been exploited for a range of malicious purposes, from phishing attacks \cite{Afane2024} and misinformation \cite{kreps2022all} to politically motivated social botnets and astroturfing campaigns \cite{woolley2016automating, keller2020political}. 
With LLMs becoming more accessible and capable, we can expect their use in criminal activity to increase as well. Malicious bots are in a constant game of cat and mouse against existing content moderation tools. This adaptation of language used to avoid algorithmic detection is widely referred to as Algospeak \cite{Steen2023Algospeak}. Steen et al. define Algospeak as a communicative practice developed in direct response to online content moderation. These behaviors can range from minor lexical substitutions to complex syntactic and semantic manipulations, reflecting an adaptive co-evolutionary dynamic between content generators and moderators. The goal of Algospeak is not to build a hidden community, but to be understood by a broad audience without algorithmic intervention by the platform’s content moderation algorithms \cite{Steen2023Algospeak}. This represents a novel aspect in human communication, as it is intended to exclude machines rather than humans. This contrasts with traditional encrypted languages, which are designed to prevent understanding by other human subgroups, even when supported by machines capable of deciphering the language.

The same goal now applies to malicious online LLM-based agents. When spreading toxic, criminal, or misleading information, they aim to maximize the reach of their message by increasing the proportion of people who can understand it regardless of background knowledge. Therefore, these agents seek to maximize understandability while minimizing the chance of detection by moderation instruments.

The phenomenon of Algospeak illustrates that this creates a continuous spiral of language modulation aimed at avoiding detection. This research provides a novel formalization of this dynamic. It introduces the concept of Majority Understandable Modulation (MUM), defined as the modulation level where additional evasive changes improve detector evasion but begin to lose comprehension for the majority of recipients. Importantly, this point is not fixed but moves depending on context shared between participants of a conversation, these can be humans or LLMs. The current literature lacks a principled understanding to quantifying the tradeoff between detectability by automated systems and comprehensibility.
The main contributions are:
\begin{enumerate}
\item A formal definition of Algospeak and key underlying dynamics at play.
\item A reference dataset, see Figure \ref{fig:modulated_ex}, and framework for modulated-dataset construction. 
\item A unified experimental framework and empirical analysis of LLMs as both interpreters (meaning recovery) and classifiers (policy-violation detection), revealing consistent MUM points.
\end{enumerate}

By establishing a measurable boundary for effective yet interpretable evasive language, our work provides a foundation for designing more robust moderation systems and understanding the co-evolutionary dynamics of language in LLM-mediated ecosystems. 

\section{Related Work}
\label{Related}


As Algospeak is a relatively new phenomenon it was first brought to attention of research through public news outlets \cite{Curtis_2022,Delkic_2022,Titz_Lehmann_2023} and more formally by \cite{Steen2023,klug2023}. \cite{Steen2023} collected 70 examples of Algospeak through semi-structured interviews with content creators. They analyzed the usage of Algospeak and the connection to TikTok’s content moderation practices. On a more structural  side, \cite{fillies2024simple} created a taxonomy for Algospeak, based on the examples by \cite{Steen2023}.

Coded language, particularly Code-Mixing and Code-Switching, has been widely studied in linguistics \cite{bali-etal-2014-borrowing}. This is especially true in the context of hate speech detection, where coded language has been thoroughly examined \cite{Barman2014,mathur-etal-2018-detecting,bohra-etal-2018-dataset}. Research in this area has primarily focused on mixed-code hate speech and its translation \cite{Tundis2020}. In the fields of Leetspeak and propaganda detection, \cite{Tundis2020} developed a supervised network to classify texts using Leetspeak encoding directly. Similarly, in image analysis, \cite{velez2023deobfuscating} employed Neural Networks to decode Leetspeak. While adversarial NLP explores trade-offs between classifier evasion and semantic preservation~\citep{garg2020bae,zhou2024humanizing}, Algospeak differs as an organic, community-driven phenomenon relying on shared context~\citep{clark1991brennan}. Our MUM metric formalizes this context-dependent comprehensibility threshold.


\section{Methodology}
\label{meth}

Clark's (1996) \cite{clark1996using} joint action model conceptualizes language use as collaborative activity requiring coordination between participants who share common ground, the mutual knowledge, beliefs, and assumptions necessary for successful communication. Three key principles from this directly inform our analysis of Algospeak:


First, Clark's distinction between addressees and bystanders \cite{clark1996using}. Linguistic analysis focuses on communication between human participants, but Algospeak introduces a novel configuration: speakers deliberately craft messages to be understood by human addressees while remaining opaque to algorithmic bystanders. This represents an inversion of classical cryptography, which seeks to exclude human eavesdroppers while remaining machine-processable.

Second, the principle of least collaborative effort \cite{clark1986referring} suggests that actors minimize joint effort in achieving mutual understanding. In the context of Algospeak, toxic actors face a trade-off: excessive modulation increases the collaborative effort required from addressees to recover meaning, potentially reducing message reach, while insufficient modulation fails to evade the content moderation bystander.

Third, the accumulation of common ground through repeated interaction \cite{clark1991brennan} explains why IMUM and MUM thresholds are context-dependent. Communities with shared cultural references, in-group terminology, or prolonged exposure to specific modulation strategies can sustain higher levels of linguistic distortion while maintaining comprehension, effectively shifting the sigmoid curve rightward along the modulation axis.


Drawing on this theoretical model of language use we define Algospeak as: A context-sensitive, multimodal register in which speakers deliberately modulate the surface form of their expressions - orthographic, lexical, morphological, phonetic, pragmatic, syntactic, or semiotic - to keep meaning and function recoverable to addressees while lowering the likelihood that fully automated moderation systems will detect that meaning. This modulation can take multiple forms, vary in intensity, and often is layered.

For a deeper understanding of the problem, three general observations and two propositions are made. The two propositions will be tested experimentally through simulation. We begin with the following core assumptions for Algospeak.

\textbf{Bystander Assumption}. The malicious actor is aware of algorithmic content moderation systems and deliberately distorts their language using methods such as Algospeak to evade detection \cite{Steen2023Algospeak,clark1996using}.

\textbf{Goal Assumption}. The objective of a toxic actor on a large social media platform is to maximize the reach of their message online; the more people who understand the intended meaning, the greater its reach \cite{pancer2019readability}.

\textbf{Common Ground Assumption}. Participants in an online conversation share a degree of common ground shaped by factors such as cultural identity and prior conversational context \cite{clark1991brennan}.


The two main propositions are: First, as linguistic modulation used in Algospeak increases, the proportion of participants able to understand the underlying meaning decreases. Depending on the degree of shared common ground, there exists a threshold of language modulation beyond which it becomes impossible for the majority of participants to grasp the intended meaning. Second, as the language becomes more modulated from its original form, it becomes increasingly difficult for trained classifiers to detect the underlying meaning and correctly classify the content of the message.

\begin{figure*}
\centering
\includegraphics[width=3.5in]{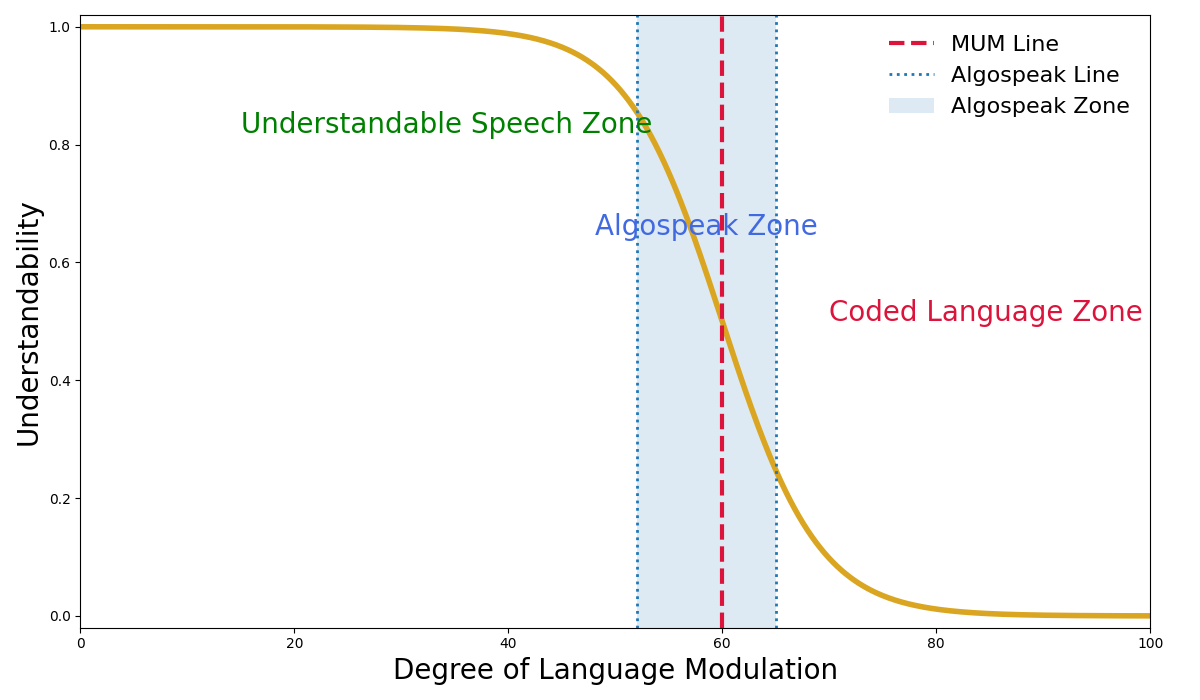}
\caption{Overview of the relationship between modulation and understandability. }
\label{relationship}
\end{figure*}


Based on these assumptions, a relationship between modulation and understandability can be proposed (see Figure \ref{iniatal_idea}, Appendix \ref{algoclass}), where understandability is defined as the percentage of participants who can recover the original meaning at a given modulation level. The four resulting zones range from typical language use (low modulation, high understandability) through opaque (low modulation, low understandability) to coded language (high modulation, low understandability) to Algospeak (high modulation, high understandability), the region most attractive to malicious actors.



Based on Wichmann and Hill \cite{wichmann2001psychometric}, we propose that the relationship between understandability and modulation (and detectability/modulation) can be modeled using a sigmoid function, see Figure \ref{relationship}. The figure illustrates that as modulation increases, the percentage of the conversation participants able to comprehend the content decreases. It further shows that beyond a certain level of modulation, the language transitions from the zone of general understandability into the Algospeak zone, and eventually into coded language. The transition from Algospeak to coded language is what we define as the Majority Understandable Modulation (MUM) point, where the majority of the population within the context can no longer follow the majority of the conversation. This means that for a population the MUM point is based on the points that individuals fail to understand most of the modulated content. We call this second point the Individual Majority Understandable Modulation (IMUM) Point. It is important to note that both the slope and the modulation level of these points are highly dependent on the shared common context between the participants, and the figure presents example values solely for illustrative purposes. These constructs are proposed for humans and LLMs alike. A formalization of the dynamics can be found in Appendix \ref{formalized}.


\begin{figure*}[t]
    \centering
    \begin{subfigure}{0.49\textwidth}
        \includegraphics[width=\linewidth]{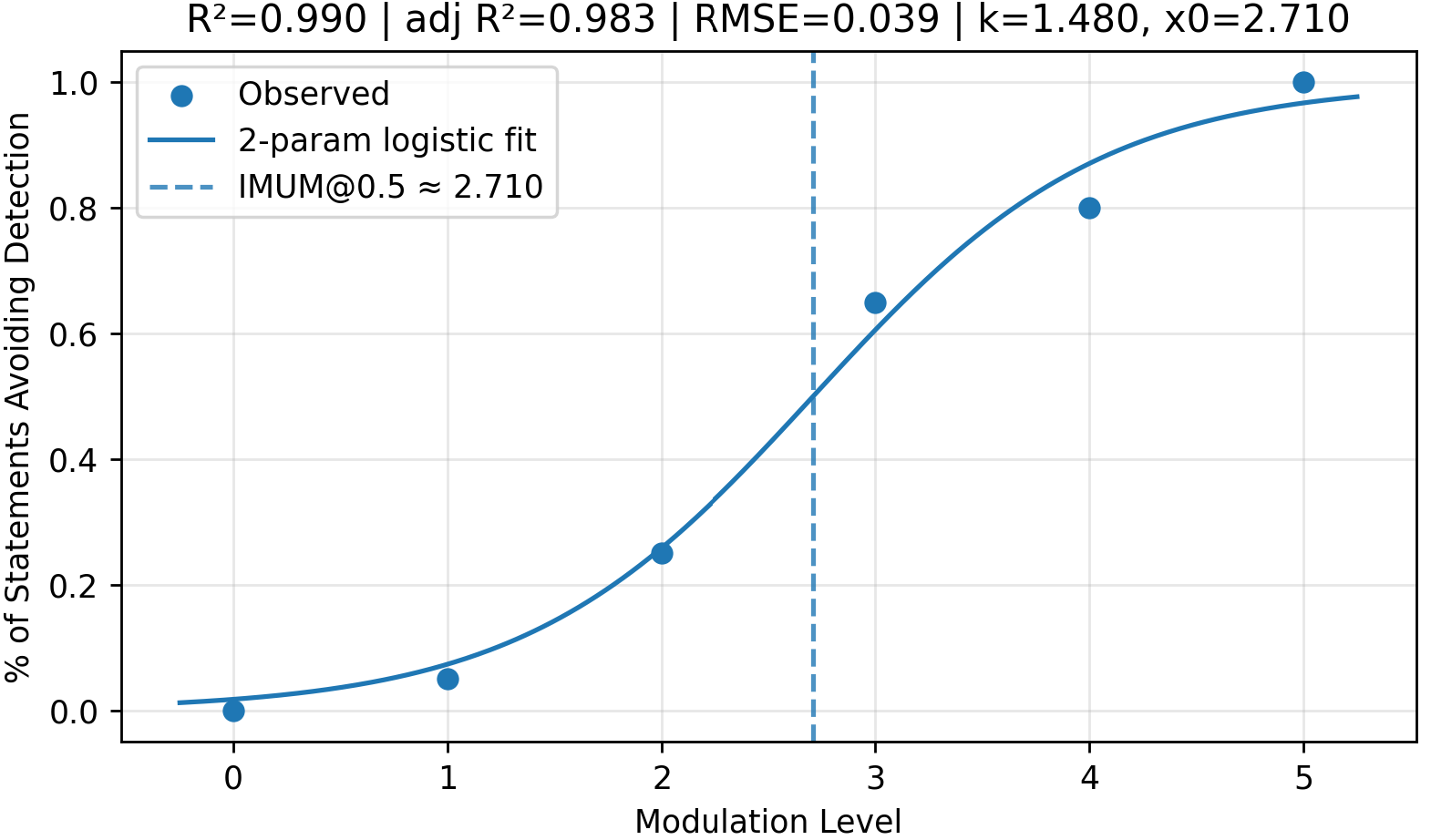}
        \caption{Well fitted curve for the code word class, detection experiment.}
        \label{fig:code_word_A}
    \end{subfigure}
    \begin{subfigure}{0.49\textwidth}
        \includegraphics[width=\linewidth]{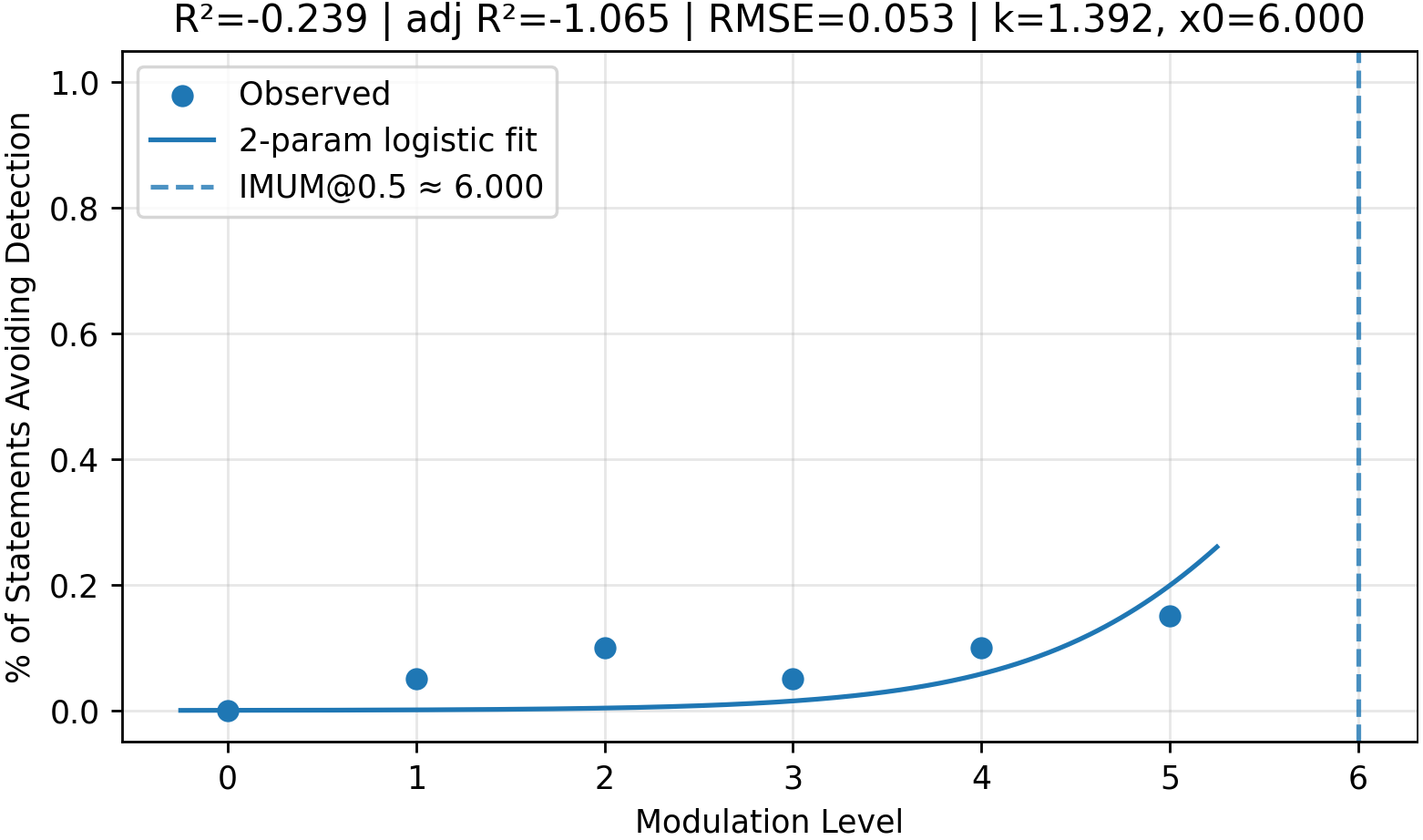}
        \caption{Ill fitted curve for the change in spelling class, detection experiment.}
        \label{fig:spelling_B}
    \end{subfigure}

    \begin{subfigure}{0.49\textwidth}
        \includegraphics[width=\linewidth]{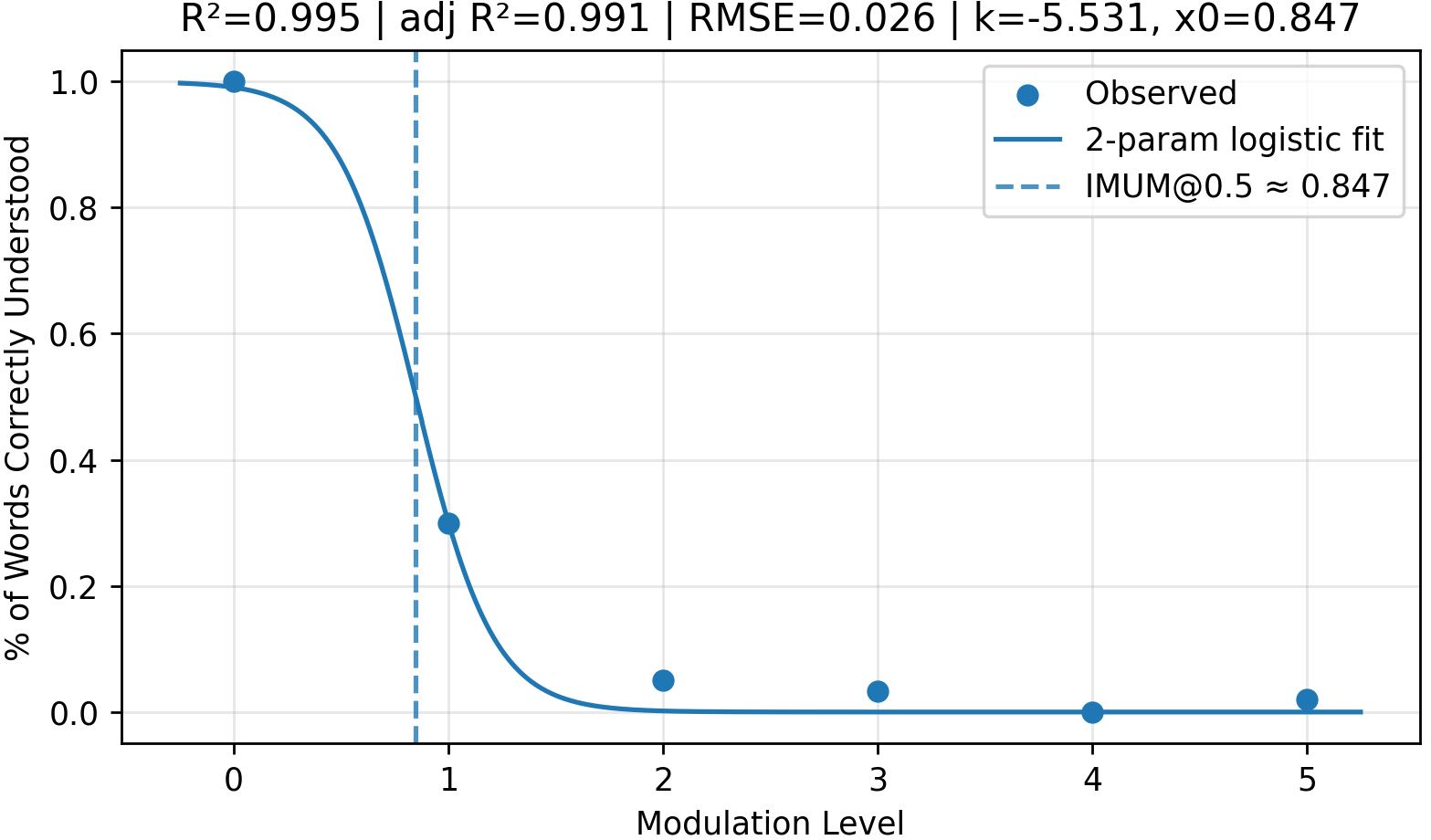}
        \caption{Well fitted curve for the code word class, understanding experiment.}
        \label{fig:code_word_C}
    \end{subfigure}
    \begin{subfigure}{0.49\textwidth}
        \includegraphics[width=\linewidth]{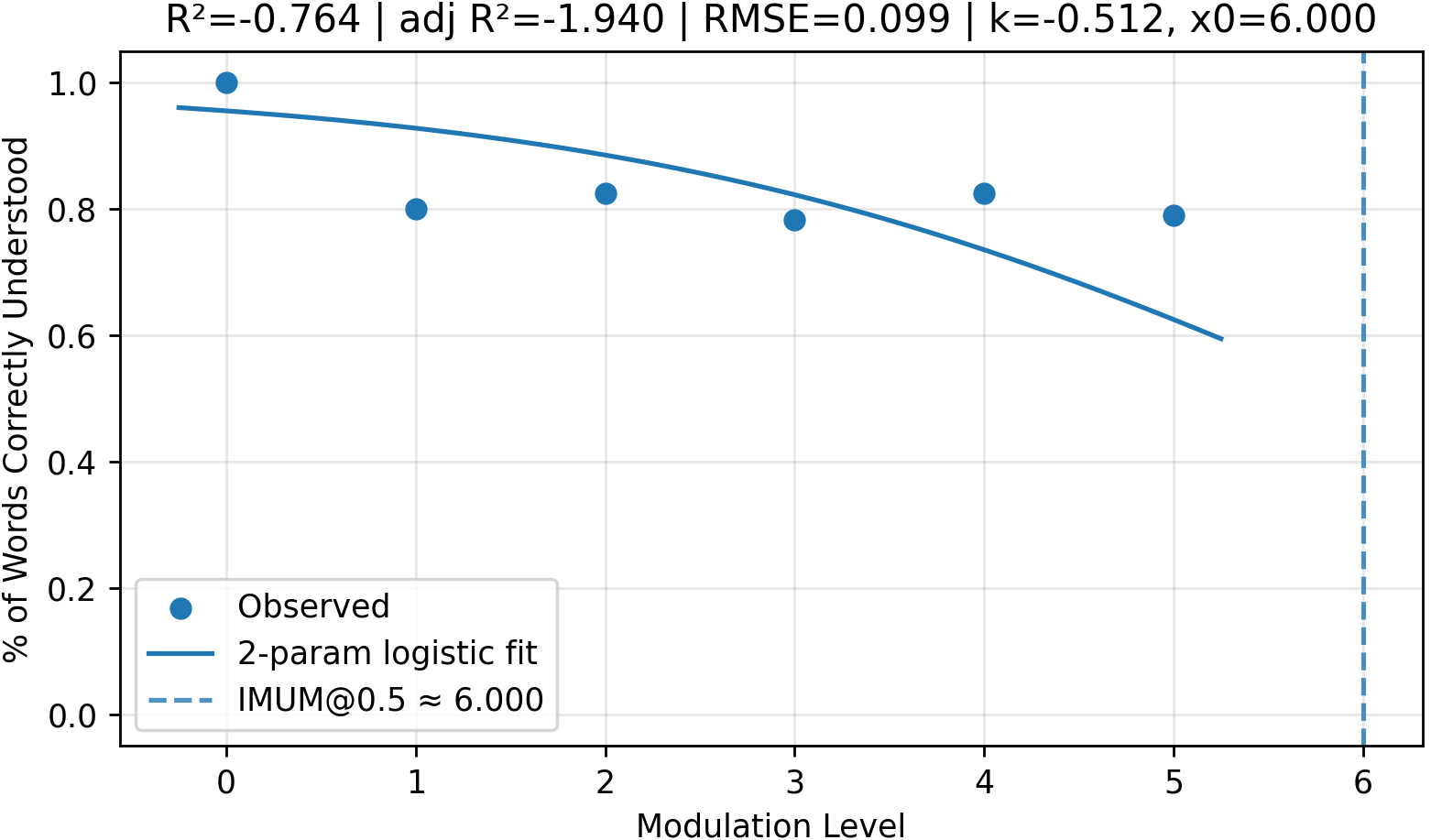}
        \caption{Ill fitted curve for the change in spelling class, understanding experiment.}
        \label{fig:spelling_C}
    \end{subfigure}
    \label{fig:three_images}
    \caption{Example results for two of the seven Algospeak strategies for the model GPT-4o-mini. }
\end{figure*}

\section{Experiments}
\subsection{Dataset Creation}

Because Algospeak is not detectable by current classifiers and no large-scale dataset exists, this study creates its own as a proof of concept. COVID-19 misinformation was chosen as the example domain due to its well-documented use of Algospeak, existing datasets, and the ease of determining ground truth, which reduces annotator bias. This focused validation allows us to test our theoretical framework under controlled conditions before extending to other domains in future work.

The dataset was created in four phases. First, we created 20 example sentences, each 10–15 words long and centered on common COVID-19 misinformation. Second, the sentences were validated to ensure the baseline model (GPT-4o) correctly classified them as misinformation, based on majority agreement across three identical trials, with identical temperature (0) and prompt settings.
Next, the model identified feature importance by selecting the six words most responsible for each misinformation classification, determined by majority vote across three identical trials. A human researcher incrementally modified each sentence, starting with the most influential words and moving to the least, to create varying modulation levels.
This process was applied to all 20 sentences across five modulation levels ($\sim$10\%, $\sim$20\%, $\sim$30\%, $\sim$40\%, and $\sim$50\%) and for all seven Algospeak strategies identified by \cite{fillies2024simple} (including altered spellings (to unknown and known words), abbreviations, pictorial representations, paraphrasing, repurposed words, and phonetic substitutions), resulting in a total of 700 modulated messages. An example of the five different modulation levels can be seen in Figure~\ref{fig:modulated_ex}.

We acknowledge that this dataset construction has limitations; see Section \ref{limi}.  We explicitly situate the work as a proof of concept and planned expansion to larger, more diverse datasets.

\subsection{Detection Experiment}
For the detection of misinformation, a prompt-based setup was created. The prompt was designed in line with prior research that employs LLMs for misinformation detection. It instructed the LLM to determine whether the given statement contained misinformation, without adding judgments or other assumptions beyond that task. The model was asked to make its decision solely based on the provided statement and the general knowledge about COVID-19 contained within its training data.

This process was repeated three times for each of the five modulation levels, and a majority score was calculated under the same temperature (0) and prompt settings.

\subsection{Understandability Experiment}
While LLMs are not capable of truly understanding the meaning of words, we operationalized understandability by having the model reconstruct each modulated word, repeating that process at every modulation level. Evaluation was conducted by calculating a similarity score between the original words and the reconstructed words, using a chosen threshold of 95\%. Statements exceeding this threshold were considered correctly identified.



\subsection{Statistical Analysis}

To characterize the relationship between distortion level and model performance, we employed two complementary approaches:
We fitted a two-parameter logistic function to model detection and understanding rates across distortion levels, where k controls the steepness of decline and x0 represents the inflection point (the Majority Understandable Modulation threshold at 50\%). Goodness of fit was assessed using R², adjusted R², and RMSE. This parameterization enables direct comparison of MUM thresholds across strategies and models.

To test whether performance declines monotonically with increasing distortion, independent of the specific sigmoid parameterization, we computed Spearman's rank correlation ($\rho$) between distortion level and performance for each series. Spearman's test is appropriate for our sample size (N=6 distortion levels) and provides robust inference without assuming a parametric functional form. Statistical significance was assessed at $\alpha$ = 0.05.

\section{Results}

We begin with a detailed analysis of a representative model (GPT-4o-mini) to illustrate how to interpret the metrics and plots, then present aggregated results for all LLMs highlighting key contrasts.

\subsection{Results for Model GPT-4o-mini}
\label{gpt-mini}
\begin{table*}[t]
\centering
\caption{Detection Results - 2-parameter logistic curve and Spearman rank correlation, GPT-4o-mini}
\begin{tabular}{llccccccc}
\toprule
\textbf{Series} & {$k$} & IMUM & {$R^2$} & {Adj. $R^2$} & {RMSE} & Sig. Fit& $\rho$ & p-val. \\ 
\midrule

Unkn. word & 1.3918 & 6.0000 & -0.2388 & -1.0646 & 0.0533 & Poor &  0.853  &   0.0416\\ 
New word  & 1.3657 & 4.1783 &  0.9959 &  0.9932 & 0.0173 & Strong &  0.986     &0.0083 \\ 
Abbreviations & 1.2885 & 4.3299 &  0.9994 &  0.9990 & 0.0064 & Strong & 0.986&     0.0139\\ 
Emoticons  & 1.4568 & 3.7796 &  0.9947 &  0.9912 & 0.0236 & Strong & 0.986     &0.0111 \\ 
Paraphrasing  & 0.8241 & 4.4341 &  0.9704 &  0.9506 & 0.0381 & Strong & 0.986  &  0.0083\\ 
Code  & 1.4799 & 2.7102 &  0.9896 &  0.9827 & 0.0388 & Strong & 1.000     &0.0055\\ 
Phonetic  & 0.7904 & 5.1973 &  0.9896 &  0.9826 & 0.0163 & Strong &0.986 &    0.0083\\ 
\bottomrule
\label{Detection_anal}
\end{tabular}
\end{table*}



\begin{table*}[htbp]
\centering
\caption{Understanding Results - 2-parameter logistic curve and Spearman rank correlation, GPT-4o-mini}
\begin{tabular}{llccccccc}
\toprule
\textbf{Series} & {$k$} & IMUM & {$R^2$} & {Adj. $R^2$} & {RMSE} & Sig. Fit &  $\rho$ & p-val.  \\ 
\midrule
Unkn. word & -0.5119 & 6.0000 & -0.7638 & -1.9397 & 0.0990 & Poor & -0.551    & 0.2667\\ 
New word& -0.3401 & 4.5889 &  0.6747 &  0.4579 & 0.0911 & Moderate & -1.000 &    0.0055 \\ 
Abbreviations & -0.4324 & 4.0659 &  0.8153 &  0.6922 & 0.0786 & Moderate & -0.943&     0.0139\\ 
Emoticons & -0.6704 & 1.4562 &  0.6536 &  0.4227 & 0.1618 & Moderate & -0.943 &    0.0194\\ 
Paraphrasing & -4.4236 & 0.7645 &  0.8024 &  0.6706 & 0.1384 & Moderate & -0.928  &   0.0139\\ 
Code & -5.5315 & 0.8475 &  0.9949 &  0.9915 & 0.0256 & Strong & -0.943    & 0.0111\\ 
Phonetic & -0.5622 & 6.0000 & -0.1454 & -0.9090 & 0.0783 & Poor & -0.928  &   0.0277 \\ 
\bottomrule
\label{tab_under}
\end{tabular}
\end{table*}


\textbf{Detection Experiment}
For each statement and Algospeak strategy, we applied majority voting at every modulation level and aggregated results across the 20 examples. This revealed clear patterns: for the code-word strategy (Figure \ref{fig:code_word_A}), replacing the most important words up to modulation level 4 caused 80\% of statements to evade misinformation classification. In contrast, for the unknown-spelling strategy (Figure \ref{fig:spelling_B}), over 80\% are still correctly flagged at modulation level 5.

To test whether detection rates decline systematically with distortion, we fitted two-parameter logistic curves and computed Spearman rank correlations. Results are shown in Table \ref{Detection_anal}. Six of seven strategies achieved strong sigmoid fits (adj. R² > 0.90), with code words (R² = 0.990) and new words (R² = 0.996) showing near-perfect fits. Only the unknown-word strategy fit poorly (adj. R² = -1.065), indicating highly variable detection even at low distortion levels.

The Spearman's test confirmed that all seven strategies exhibited significant monotonic decreases in detection (p < 0.05), with six achieving near-perfect correlation ($\rho$>= 0.98). This validates that increased distortion genuinely reduces detection across all strategies.

The IMUM for each strategy indicates how quickly it distorts text and enables direct comparison. For instance, on average, introducing just 2.7 code words caused 50\% of statements to be misclassified, compared with 4.4 paraphrased words needed for the same effect.




\textbf{Understandability Experiment}
Reconstructions were repeated for each modulated word at every modulation level and averaged across 20 examples for all seven strategies. As Figure \ref{fig:code_word_C} shows, introducing a single unknown code word allowed correct reconstruction in only 40\% of cases. In contrast, strategies preserving surface form, like spelling or phonetic changes (Figure \ref{fig:spelling_C}), enabled correct reconstruction for most terms at all levels.


Similar to the detection experiment, we computed Spearman correlations to test whether understanding declines with distortion. Results are shown in Table \ref{tab_under}. Three strategies achieved strong or moderate fits (code words adj. R² = 0.99, paraphrasing 0.67, abbreviations 0.69), while unknown-word and phonetic strategies fit poorly (adj. R² < 0). Six of seven strategies showed significant monotonic decreases in understanding (p < 0.05), with three achieving perfect negative rank correlation ($\rho$ = -1.0). Only the unknown-word strategy failed to reach significance ($\rho$ = -0.551, p = 0.2667).

Code words required only 0.85 modulations to cross 50\% understanding, while strategies preserving surface form (phonetic, spelling) never crossed this threshold within our tested range (x0 = 6.0).

\begin{table*}[htbp]
    \centering
    \caption{Detection Results: Adjusted $R^2$ with  majority fit estimation and Spearman Rank Correlation Significance by Strategy and Model}
    \label{tab:detection_combined}
    \small
    \begin{tabular}{@{}lccccccccc@{}}
        \hline
        Strategy & Claude & GPT-4o-m & GPT-4o & Llama & Mistral & Qwen & Grok & Maj. & Sig. C.\\
        \hline
        Unkn. word     & -4.83\,\texttimes & -2.03\,\checkmark & -0.69\,\checkmark & -0.18\,\checkmark &  0.70\,\checkmark &  0.89\,\checkmark & -3.17\,\texttimes & P & 5/7\\
        New word       &  0.90\,\checkmark &  0.99\,\checkmark &  0.99\,\checkmark &  0.90\,\checkmark &  0.77\,\checkmark &  0.96\,\checkmark &  0.93\,\checkmark & S & 7/7\\
        Abbreviation   &  0.92\,\checkmark &  0.98\,\checkmark &  0.95\,\checkmark &  0.96\,\checkmark & -0.05\,\checkmark &  0.91\,\checkmark &  0.98\,\checkmark & S & 7/7 \\
        Emotion        & -8.92\,\texttimes &  0.98\,\checkmark &  0.96\,\checkmark &  0.95\,\checkmark &  0.88\,\checkmark &  0.99\,\checkmark &  0.94\,\checkmark & S & 6/7\\
        Paraphrase     &  0.99\,\checkmark &  0.95\,\checkmark &  0.93\,\checkmark &  0.28\,\checkmark & -4.88\,\checkmark &  0.95\,\checkmark &  0.97\,\checkmark & S & 7/7\\
        Code           &  0.91\,\checkmark &  0.96\,\checkmark &  1.00\,\checkmark &  0.95\,\checkmark &  0.96\,\checkmark &  0.97\,\checkmark &  0.99\,\checkmark & S & 7/7\\
        Phonetic       &  0.30\,\checkmark &  0.99\,\checkmark & -4.54\,\checkmark & -0.25\,\checkmark &  0.12\,\texttimes &  0.89\,\checkmark &  0.35\,\checkmark & P & 6/7\\
        \hline
        Total          & 5/7 & 7/7 & 7/7 & 7/7 & 6/7 & 7/7 & 6/7 &  & 86\% \\
        \hline
    \end{tabular}
    \vspace{2mm}
    
    \footnotesize{Adjusted $R^2$ values shown with significance: \checkmark\ = $p < 0.05$, \texttimes\ = $p \geq 0.05$; S = Strong, M = Moderate, P = Poor}
\end{table*}

\begin{table*}[htbp]
    \centering
    \caption{Understandability Results: Adjusted $R^2$ with  majority fit estimation and Spearman Rank Correlation Significance by Strategy and Model}
    \label{tab:understanding_combined}
    \small
    \begin{tabular}{@{}lccccccccc@{}}
        \hline
        Strategy & Claude & GPT-4o-m & GPT-4o & Llama & Mistral & Qwen & Grok & Maj. & Sig. C.\\
        \hline
        Unkn. word   & -1.03\,\checkmark & -1.94\,\texttimes & -2.46\,\texttimes &  0.34\,\texttimes &  0.49\,\texttimes &  0.47\,\texttimes & -2.39\,\texttimes & P & 1/7 \\
        New word     &  0.90\,\checkmark &  0.46\,\checkmark &  0.67\,\checkmark &  0.81\,\checkmark &  0.87\,\checkmark &  0.87\,\checkmark &  0.35\,\checkmark & S & 7/7 \\
        Abbreviation &  0.88\,\checkmark &  0.69\,\checkmark &  0.23\,\checkmark &  0.76\,\checkmark &  0.36\,\texttimes &  0.38\,\texttimes & -0.34\,\checkmark & P & 5/7\\
        Emotion      &  0.55\,\checkmark &  0.42\,\checkmark &  0.25\,\checkmark &  0.85\,\texttimes &  0.74\,\texttimes &  0.74\,\texttimes &  0.51\,\checkmark & M & 4/7\\
        Paraphrase   &  0.63\,\checkmark &  0.67\,\checkmark &  0.67\,\texttimes &  0.98\,\checkmark &  1.00\,\checkmark &  1.00\,\checkmark &  0.54\,\checkmark & M & 6/7\\
        Code         &  0.86\,\checkmark &  0.99\,\checkmark &  0.76\,\checkmark &  1.00\,\checkmark &  1.00\,\texttimes &  1.00\,\texttimes &  0.83\,\checkmark & S & 5/7 \\
        Phonetic     & -0.57\,\checkmark & -0.91\,\checkmark & -1.96\,\checkmark &  0.11\,\checkmark &  0.38\,\checkmark &  0.39\,\checkmark & -7.45\,\texttimes & P & 6/7\\
        \hline
        Total        & 7/7 & 6/7 & 5/7 & 5/7 & 3/7 & 3/7 & 5/7 & & 67\% \\
        \hline
    \end{tabular}
    \vspace{2mm}
    
    \footnotesize{Adjusted $R^2$ values shown with significance: \checkmark\ = $p < 0.05$, \texttimes\ = $p \geq 0.05$; S = Strong, M = Moderate, P = Poor}
\end{table*}

\subsection{Cross-model comparison}

We compared seven state-of-the-art models, ranging from open- to closed-source and varying in the scale of their training parameters. The models considered were: claude-sonnet-4-5, gpt-4o-mini, gpt-4o, llama-3.1-8b-instant, Ministral-8B-Instruct-2410, Qwen3-VL-32B-Instruct-FP8, grok-4.1-fast.

\textbf{Detection performance across models:}
Similar to Section ~\ref{gpt-mini}, we tested whether increased distortion reduces detection using both sigmoid curve fitting (for MUM estimation) and Spearman's rank correlation (for statistical inference). Table \ref{tab:detection_combined} reports goodness-of-fit (adjusted R²) for the sigmoid curves and Spearman correlation results across all models. Heatmaps Figure \ref{fig:heat_detection} in Appendix \ref{heatmap} reports adjusted $R^2$ values by strategy for the seven models\footnote{Adjusted $R^2$ can be negative (and unbounded below), strongly negative values indicate a fit worse than a constant baseline.} As before, most strategies fit the sigmoid curve well, whereas the change-in-spelling strategy producing an unknown word continues to show poor fit. Interestingly, under majority voting, most models also did not seem to be strongly affected by the introduction of phonetic resemblance, an issue already indicated by the performance of the understanding task in Section \ref{gpt-mini}. 

Spearman's test revealed that 86\% (42/49) of (model, modulation class) pairs exhibited statistically significant monotonic decreases in detection (p < 0.05), with 43\% showing perfect or near-perfect rank correlation (|$\rho$| >= 0.98). This confirms that the relationship between modulation and detection represents genuine monotonic trends robust across models and strategies.

Overall, the results suggest that the choice of strategy matters more than the choice of model, although some models appear to work well with specific strategies.

\textbf{Understandability across models:}
Similar to Section ~\ref{gpt-mini}, we tested whether understanding declines with distortion using sigmoid fitting and Spearman's test. Table \ref{tab:understanding_combined} reports adjusted R² values and Spearman significance results.

The heatmaps Figure \ref{fig:heat_dunders} in Appendix \ref{heatmap} report adjusted $R^2$ values by strategy for the seven models, along with the majority measure of goodness-of-fit. As before, most strategies show a moderate to strong sigmoid fit, while the change-in-spelling and phonetic strategies perform poorly. Moreover, although Claude, GPT-4o-mini, and Llama exhibit strong or moderate fit for the abbreviation strategy, weaker fits for the remaining models yield an overall weak fit, highlighting substantial cross-model differences within a strategy.

Spearman's test revealed that 67\% (33/49) of (model, distortion-class) pairs exhibited statistically significant monotonic decreases in understanding (p < 0.05). This lower rate compared to detection (86\%) suggests that understanding is more difficult to operationalize and varies more across models. Claude achieved 100\% significant trends (7/7), while Mistral and Qwen managed only 43\% (3/7), revealing substantial model-specific differences in semantic reconstruction capabilities.



\textbf{Trade-off / IMUM \& MUM comparison}
When combining the individual IMUM values for each strategy and model, we obtain an overall MUM per strategy that captures the modulation level beyond which most models fail to understand the majority of the text. Figure~\ref{fig:llm_compare} shows both the model-specific IMUM values and the overall MUM for the code-word-introduction strategy. 

\section{Discussion}
The experiments confirm both propositions across most strategies and models.

The results of the first experiment show that for most of the seven modulation strategies and LLMs, the proposition holds: as modulation increases, classifier detection performance decreases, and the proposed sigmoid function fits mostly well.

It is noteworthy that the strategies that don't follow this trend are common and straightforward. They consist of phonetic resemblance, without creating a new word, such as "Covid = Kovit", and the spelling alteration by replacing certain characters with others (for example, numbers). This outcome was expected, as phonetic resemblance is close in nature to spelling mistakes and spelling alterations rely heavily on well-established substitution rules such as “@ = a” or “3 = e.” These conventions are widely used and recognized,  therefore do not pose a challenge for LLM-based detection.


These inter-strategy differences have further implications. At the strategy level, our results suggest that shifting toxic communication from dynamic, unstructured Algospeak strategies (e.g., paraphrasing) to more formal Algospeak strategies closer to coded language (e.g., abbreviations) is beneficial for moderation, as detection capabilities increase while general comprehensibility decreases. Such a shift raises the time and cognitive costs for toxic actors, who must invest effort into developing a structured coded language and communicating its meaning to their intended audience. It also increases the initial effort required for external participants to join the conversation or be influenced by it.

The second experiment supported the propositions that increasing modulation reduces the ability to detect the original form of a word or sentence, and thus decreases overall comprehension. All strategies and LLMs showed a decline in understandability performance as modulation increased, some were affected more strongly than others. A moderate or strong fit of the proposed sigmoid curve was observed for most strategies and LLMs. Notably, some strategies disrupt detection while leaving reconstruction largely intact. This likely reflects a cue asymmetry: modulation scrambles the semantic patterns detectors rely on while preserving enough surface structure for reconstruction, possibly compounded by detectors being fine-tuned on cleaner data and thus sensitive to out-of-distribution noise. 


Overall, these results suggest that the signals used for detecting information and for understanding or reconstructing meaning are somewhat disjoint, even though they do tend to move together. Some strategies, particularly emoticons, were reliably detected (6/7 models significant) but harder to reconstruct (4/7 models), suggesting models can flag "anomalous patterns" without necessarily recovering meaning. Conversely, unknown-word substitutions proved uniquely challenging for both detection and understanding, with only 5/7 and 1/7 models achieving significance respectively. The Spearman's rank correlation provides robust statistical evidence for our core propositions.


While human judgment typically requires comprehension, LLM-based detection can succeed without it. This insight may stem from factors such as fine-tuning strategies, characteristics of the training data, safety layers, or other architectural components, highlighting a novel distinction between detection and understanding in LLMs.

\section{Conclusion and Future Work}

The research establishes a formal definition of Algospeak and outlines its underlying dynamics, supported by a proof-of-concept LLM-based experimental setup using seven large language models. It introduces a framework for creating the first five-level modulated dataset and provides a dataset comprised of 700 distinct examples across seven Algospeak categories. The findings indicate that as Algospeak increases, detection ability decreases. Furthermore, higher levels of Algospeak reduce the understandability of the modulated terms. The underlying dynamics can be partially modeled using sigmoid curves.

This work lays the groundwork for future human-subject, cross-strategy, and cross-topic experiments, and its implications are substantial, offering guidance for both moderation practices and the assessment of moderation effectiveness.

\section{Limitations}
\label{limi}

All results were generated through LLM-based settings. While human-subject experiments remain necessary to fully substantiate these findings, LLM-based agents are already active in this space, making the results directly relevant. 


This study is a proof of concept focusing on COVID-19 misinformation in English. We focus on large platforms (TikTok, Facebook, YouTube, X/Twitter) where toxic actors maximize audience size, though the framework applies to smaller platforms where reach assumptions may differ. Generalization to other domains such as hate speech or political persuasion remains untested. The dataset relies on 20 base sentences (700 total examples), which constrains semantic diversity and may introduce memorization effects. Human-authored modulations may not capture organic Algospeak variations. All experiments were English-only; modulation strategies may vary across languages. The small number of modulation levels limits statistical precision.

Despite these limitations, this work establishes a formal foundation for studying Algospeak and provides evidence for core propositions. We view this as a first step toward broader research spanning domains, languages, and human populations.



\section{Ethical Considerations}
This work is inherently dual-use: systematically identifying which Algospeak modulation strategies most effectively degrade detector performance could, if misused, provide malicious actors with a blueprint for more resilient abuse and disinformation. To mitigate this risk, we work exclusively with researcher-authored, synthetic COVID-19 misinformation statements; do not publicly release full datasets, prompts, or code, instead providing controlled access only to validated researchers with a documented safety, governance, or harm-mitigation agenda and appropriate data-handling commitments; and avoid publishing concrete high-evasion "recipes," platform-specific implementation details that would directly operationalize evasive capabilities at scale. We therefore frame this work explicitly as red-teaming in support of defense: the analyses, thresholds, and tools are intended to guide the development and evaluation of protective systems and to inform governance processes, not to be used directly in production settings or for adversarial optimization.



\appendix

\section{Classifications of language based on modulation and understandability}
\label{algoclass}
Figure \ref{iniatal_idea} depicts the classifications of language based on modulation and understandability.
\begin{figure}
\centering
\includegraphics[width=3in]{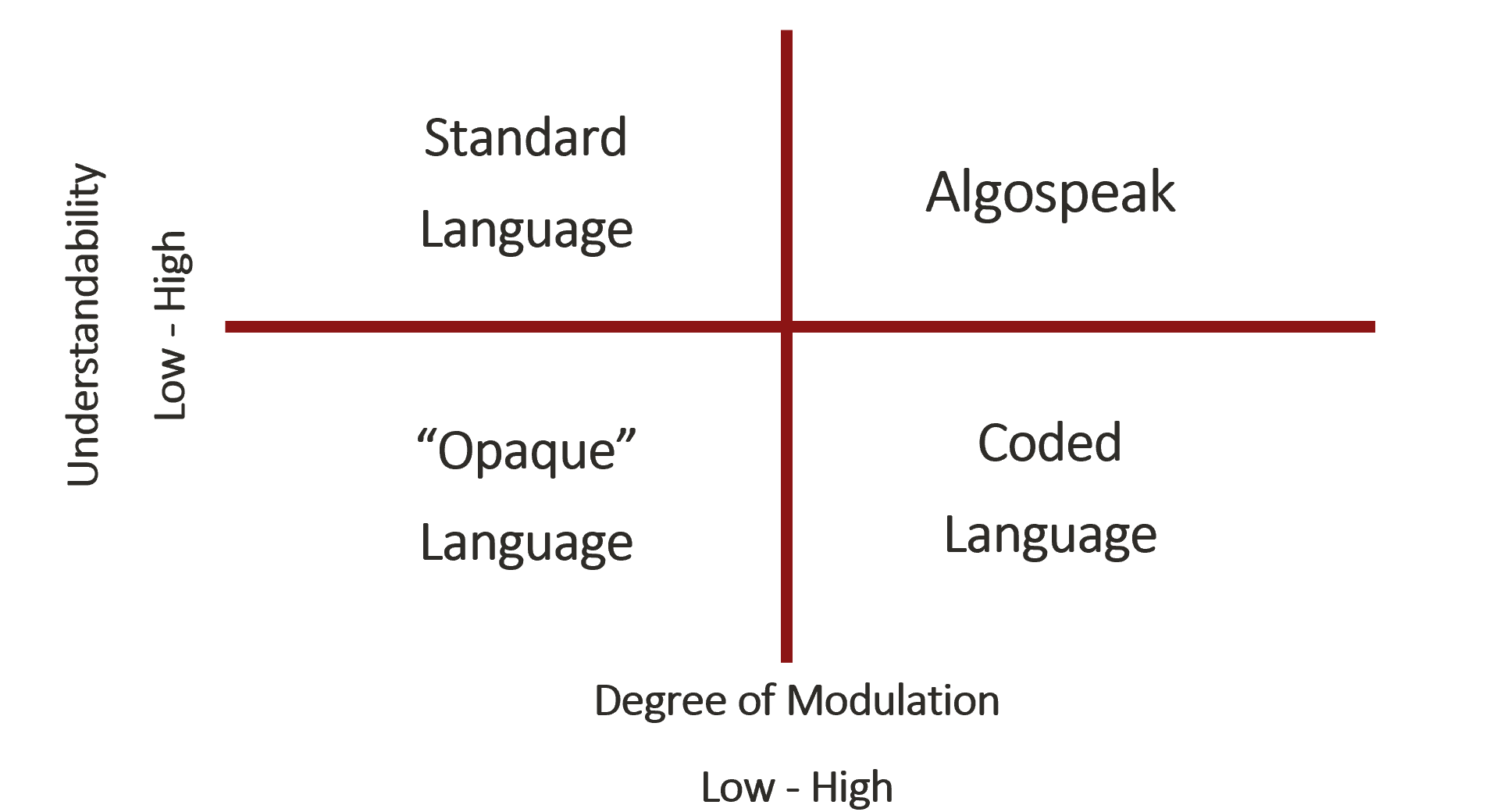}
\caption{Classifications of language based on modulation and understandability. }
\label{iniatal_idea}
\end{figure}

\section{Classifications of language based on modulation and understandability}
\label{algoclass}
Figure \ref{iniatal_idea} depicts the classifications of language based on modulation and understandability.
\begin{figure}
\centering
\includegraphics[width=3in]{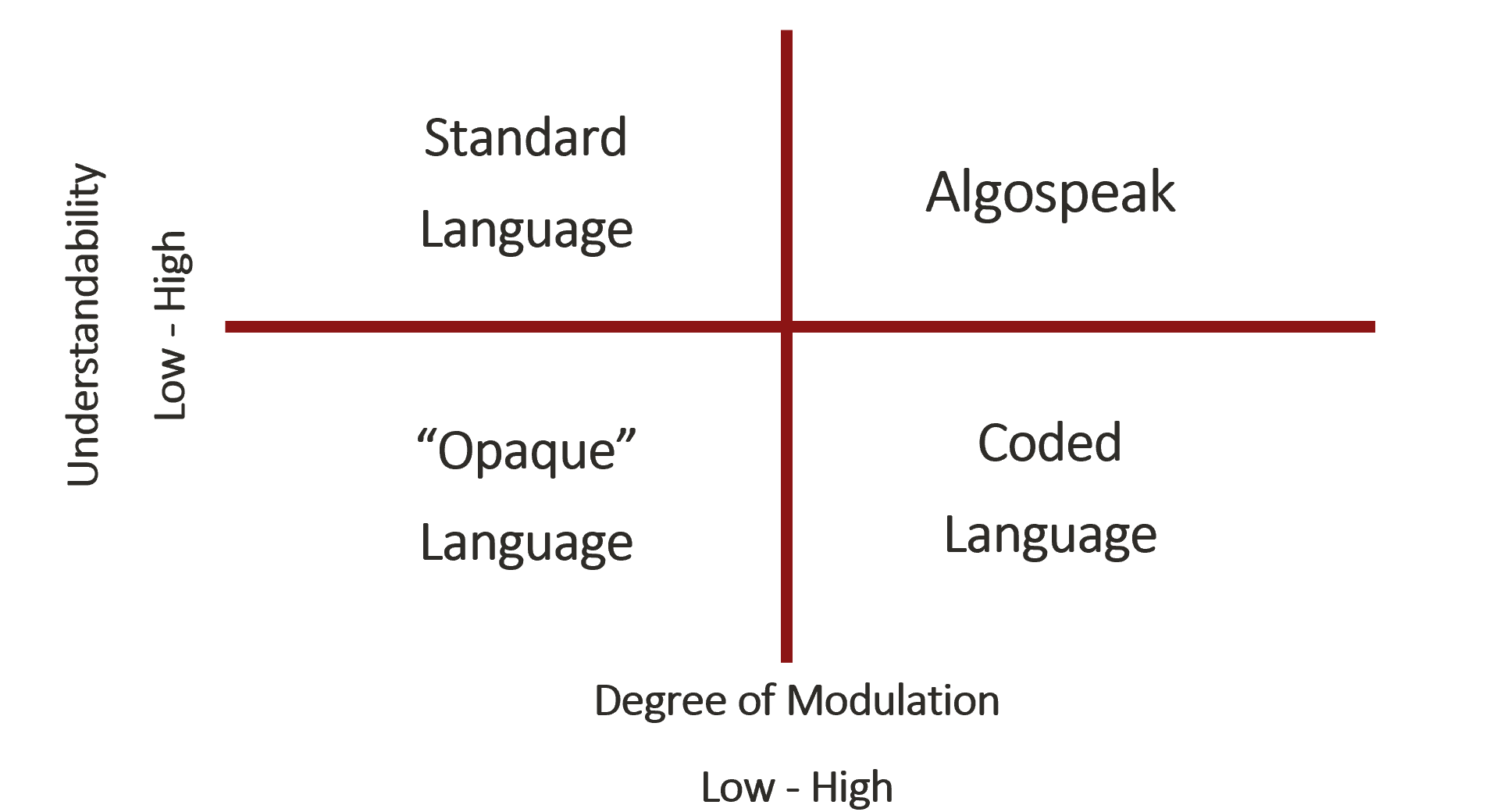}
\caption{Classifications of language based on modulation and understandability.}
\label{iniatal_idea}
\end{figure}

\section{Formalization of Majority Understandable Modulation}
\label{formalized}
Let 
\begin{itemize}
    \item $\theta \in \mathcal{X}$ denote an original text with intended meaning $c \in \mathcal{C}$,
    \item $T_d : \mathcal{X} \to \Delta(\mathcal{X})$ be a stochastic modulation operator parameterized by $d \ge 0$, representing the level of modulation, where $\Delta(\mathcal{X})$ denotes the space of probability distributions over $\mathcal{X}$.
\end{itemize}

\paragraph{Semantic Fidelity.}
A modulated text $x = T_d(\theta)$ is \emph{semantically valid} if it preserves the original meaning:
\[
S(\theta, x) =
\begin{cases}
1, & \text{if } x \text{ has the same meaning as } \theta, \\
0, & \text{otherwise.}
\end{cases}
\]

\paragraph{Understandability.}
For a population $\mathcal{H}_\kappa$ within context $\kappa$ (shared background, community, etc.), and an original text $\theta$ with meaning $c$, define
\[
\begin{aligned}
U_\kappa(d; \theta)
&= \mathbb{E}_{x \sim T_d(\theta)} \Big[ \mathbb{1}\{S(\theta,x)=1\} \cdot \Pr_{h \sim \mathcal{H}_\kappa}
\\
&\quad \big[\, h \text{ correctly infers } c \text{ from } x \,\big] \Big],
\end{aligned}
\]
where $U_\kappa(d; \theta) \in [0,1]$ measures the expected fraction of participants who can understand a modulated message sampled from $T_d(\theta)$. The indicator function $\mathbb{1}\{S(\theta,x)=1\}$ ensures that only semantically valid modulations contribute to understandability.

Based on empirical observations, $U_\kappa(d; \theta)$ can be modeled as an \emph{inverted sigmoid} decay with increasing modulation:
\[
U_\kappa(d; \theta) \approx \frac{1}{1 + e^{\alpha_\kappa (d - \beta_\kappa)}},
\]
where $\alpha_\kappa > 0$ controls the slope of decline and $\beta_\kappa$ corresponds to the inflection point at which comprehension drops below $0.5$. 
The parameter $\beta_\kappa$ captures the influence of shared common ground: larger $\beta_\kappa$ indicates higher tolerance for modulation within that context.

\paragraph{Detectability.}
For a moderation or detection model $M$,
\[
D_M(d; \theta)
= \mathbb{E}_{x \sim T_d(\theta)} 
    \Pr\!\big[\, M(x) = 1 \,\big],
\]
where $D_M(d; \theta) \in [0,1]$ denotes the probability that a modulated message sampled at distortion level $d$ is flagged by the detector.

\paragraph{Individual Majority Understandable Modulation (IMUM).}
Given a comprehension threshold $\tau \in (0,1)$ (e.g., $\tau = 0.5$ for majority understanding), the \emph{Individual Majority Understandable Modulation} for a specific text item $\theta$ is defined as
\[
\boxed{
\mathrm{IMUM}_\kappa^\tau(\theta)
= \inf \{\, d \ge 0 : U_\kappa(d; \theta) < \tau \,\}.
}
\]
That is, the IMUM is the smallest modulation level at which fewer than a fraction~$\tau$ of participants within context~$\kappa$ can correctly interpret the intended meaning of the specific text $\theta$. The term ``Individual'' refers to this being computed for an individual text item.

\paragraph{Aggregate Majority Understandable Modulation (MUM).}
To obtain an aggregate measure across a distribution of messages, let $\mathcal{D}$ be a distribution over base texts $\theta \in \mathcal{X}$. The aggregate MUM is defined as:
\[
\mathrm{MUM}_\kappa^\tau = \mathbb{E}_{\theta \sim \mathcal{D}} \big[ \mathrm{IMUM}_\kappa^\tau(\theta) \big].
\]

\paragraph{Trade-off Optimization.}
A malicious actor seeks to maximize understandability while minimizing algorithmic detectability. 
This trade-off can be formalized as the optimization problem
\[
d^*(\theta) = \arg\max_{d \ge 0} \; U_\kappa(d; \theta) \, \big(1 - D_M(d; \theta)\big),
\]
where $d^*(\theta)$ represents the optimal modulation level that balances communicative reach with moderation evasion for a given text $\theta$. 
Typically, $d^*(\theta)$ lies just below $\mathrm{IMUM}_\kappa^\tau(\theta)$, corresponding to the region commonly identified as the \emph{Algospeak zone}.

\section{Heatmaps Understandability and Detection}
\label{heatmap}
\begin{figure*}[t]
    \centering
        \includegraphics[width=\linewidth]{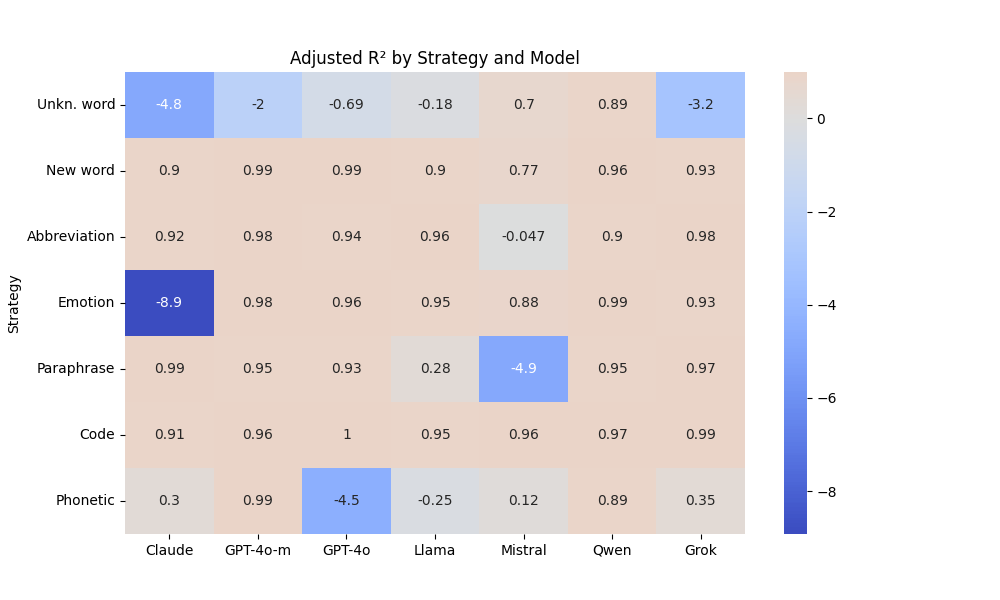}
        \caption{Heatmap detection adjusted $R^2$ by strategy and model.}
        \label{fig:heat_detection}
\end{figure*}

\begin{figure*}[t]
    \centering
        \includegraphics[width=\linewidth]{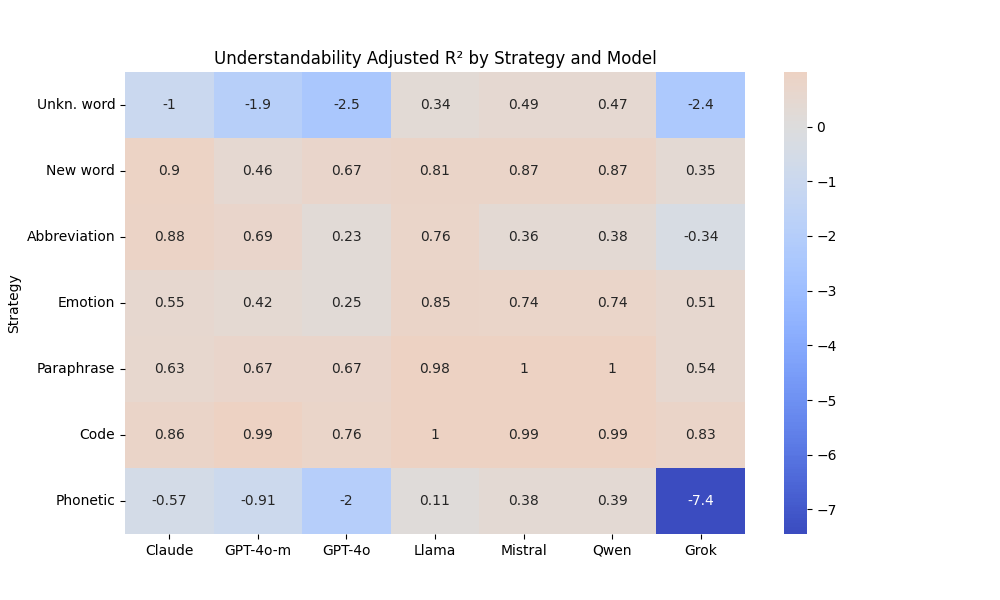}
        \caption{Heatmap understanding adjusted $R^2$ by strategy and model.}
        \label{fig:heat_dunders}

\end{figure*}

Figure \ref{fig:heat_detection} displays a heatmaps of the adjusted $R^2$ by strategy and model for the detection experiment. Figure \ref{fig:heat_dunders} displays a heatmaps of the adjusted $R^2$ by strategy and model for the understandability experiment.

\section{IMUM's Per Strategy and Model}
\label{MUM}
The tables display estimated threshold IMUM@0.5 for detection (Table \ref{tab:x0_detection_strategies_models}) and understanding (Table \ref{tab:x0_understanding_strategies_models}), both tables are by strategy and model. 

\begin{table*}[htbp]
    \centering
    \caption{Estimated threshold $x_0$ (IMUM@0.5, detection) by strategy and model}
    \label{tab:x0_detection_strategies_models}
    \begin{tabular}{lrrrrrrr}
        \hline
        Strategy & Claude & GPT-4o mini & GPT-4o & Llama & Mistral & Qwen & xAI \\
        \hline
        Unknown word & 5.09988 & 5.10000 & 5.10000 & 5.10000 & 5.10000 & 4.74839 & 5.09993 \\
        New word     & 3.28074 & 4.17831 & 3.65689 & 4.88509 & 5.09823 & 3.96294 & 2.99138 \\
        Abbreviation & 4.88880 &   4.3299   & 4.69842 & 4.57660 & 5.10000 & 3.98938 & 3.58646 \\
        Emotion      & 5.10000 & 3.81083 & 3.25907 & 3.09478 & 4.09506 & 3.35181 & 5.10000 \\
        Paraphrase   & 3.86082 & 4.43407 & 3.39559 & 5.10000 & 5.10000 & 4.34685 & 3.77906 \\
        Code         & 1.70700 & 3.69793 & 1.99989 & 2.67578 & 2.59317 & 2.23205 & 1.66515 \\
        Phonetic     & 5.10000 & 4.98789 & 5.10000 & 5.10000 & 5.10000 & 4.43758 & 5.10000 \\
        \hline
    \end{tabular}
\end{table*}

\begin{table*}[htbp]
    \centering
    \caption{Estimated threshold $x_0$ (IMUM@0.5, understanding) by strategy and model}
    \label{tab:x0_understanding_strategies_models}
    \begin{tabular}{lrrrrrrr}
        \hline
        Strategy & Claude & GPT-4o mini & GPT-4o & Llama & Mistral & Qwen & xAI \\
        \hline
        Unknown word &  6.0000 &  6.0000 & -0.9509 & 1.5788 & 0.7772 & 0.7772 & -0.8621 \\
        New word     &  6.0000 &  4.5775 &  6.0000 & 0.8103 & 0.8254 & 0.8254 &  6.0000 \\
        Abbreviation &  4.7725 &  4.0729 &  6.0000 & 1.5595 & 0.8480 & 0.8163 &  6.0000 \\
        Emotion      &  2.6879 &  1.5405 &  2.4005 & 0.7615 & 0.7615 & 0.7615 &  4.3287 \\
        Paraphrase   &  1.8680 &  0.7645 &  0.7555 & 0.7511 & 0.7863 & 0.7863 &  2.7983 \\
        Code         &  1.0062 &  0.8475 &  1.0555 & 0.7977 & 0.5005 & 0.5005 &  1.1699 \\
        Phonetic     &  6.0000 &  6.0000 &  6.0000 & 4.6037 & 3.0440 & 3.0115 & -0.9952 \\
        \hline
    \end{tabular}
\end{table*}



\bibliographystyle{unsrt}
\bibliography{lib}


\newpage
\section*{NeurIPS Paper Checklist}

\begin{enumerate}

\item {\bf Claims}
    \item[] Question: Do the main claims made in the abstract and introduction accurately reflect the paper's contributions and scope?
    \item[] Answer: \answerYes{}
    \item[] Justification: The abstract and introduction clearly state the three contributions and explicitly frame the work as a proof-of-concept study, matching the scope of the experimental results.

\item {\bf Limitations}
    \item[] Question: Does the paper discuss the limitations of the work performed by the authors?
    \item[] Answer: \answerYes{}
    \item[] Justification: A dedicated Section 8 discusses limitations including dataset size, English-only scope, synthetic modulation, and the absence of human-subject validation.

\item {\bf Theory assumptions and proofs}
    \item[] Question: For each theoretical result, does the paper provide the full set of assumptions and a complete (and correct) proof?
    \item[] Answer: \answerYes{} 
    \item[] Justification: Core assumptions are stated in Section 3 and the formal definitions and mathematical framework are provided in Appendix B.

    \item {\bf Experimental result reproducibility}
    \item[] Question: Does the paper fully disclose all the information needed to reproduce the main experimental results of the paper to the extent that it affects the main claims and/or conclusions of the paper (regardless of whether the code and data are provided or not)?
    \item[] Answer: \answerYes{} 
    \item[] Justification: Dataset construction, prompt setup, majority voting procedure, similarity threshold, and statistical analysis methods are fully described in Section 4. Material itself is only available on request by verified researchers with ongoing research projects due to dual-use concerns.

\item {\bf Open access to data and code}
    \item[] Question: Does the paper provide open access to the data and code, with sufficient instructions to faithfully reproduce the main experimental results, as described in supplemental material?
    \item[] Answer: \answerNo{} 
    \item[] Justification: For ethical reasons, full datasets and code are not publicly released. Controlled access is available to qualified researchers with a documented safety or harm-mitigation agenda, as described in Section 9.

\item {\bf Experimental setting/details}
    \item[] Question: Does the paper specify all the training and test details (e.g., data splits, hyperparameters, how they were chosen, type of optimizer) necessary to understand the results?
    \item[] Answer: \answerYes{} 
    \item[] Justification: Temperature settings, prompt design, majority voting procedure, and other aspects are specified in Section 4.

\item {\bf Experiment statistical significance}
    \item[] Question: Does the paper report error bars suitably and correctly defined or other appropriate information about the statistical significance of the experiments?
    \item[] Answer: \answerYes{} 
    \item[] Justification: Spearman rank correlation with p-values and sigmoid goodness-of-fit metrics (R², adjusted R², RMSE) are reported for all models and strategies in Section 5.

\item {\bf Experiments compute resources}
    \item[] Question: For each experiment, does the paper provide sufficient information on the computer resources (type of compute workers, memory, time of execution) needed to reproduce the experiments?
    \item[] Answer: \answerNo{} 
    \item[] Justification: Experiments relied on API calls to commercial and open-source LLMs. Exact compute costs and execution times were not reported as the scope was small, and costs are not directly comparable across closed-source models and open-source models (subject to varying API rate limits and quotas).
    
\item {\bf Code of ethics}
    \item[] Question: Does the research conducted in the paper conform, in every respect, with the NeurIPS Code of Ethics \url{https://neurips.cc/public/EthicsGuidelines}?
    \item[] Answer: \answerYes{}
    \item[] Justification: The paper conforms to the NeurIPS Code of Ethics. Ethical considerations including dual-use risks and safeguards are discussed in Section 9.

\item {\bf Broader impacts}
    \item[] Question: Does the paper discuss both potential positive societal impacts and negative societal impacts of the work performed?
    \item[] Answer: \answerYes{} 
    \item[] Justification: Section 9 discusses dual-use risks and frames the work explicitly as red-teaming in support of defensive moderation systems.
    
\item {\bf Safeguards}
    \item[] Question: Does the paper describe safeguards that have been put in place for responsible release of data or models that have a high risk for misuse (e.g., pre-trained language models, image generators, or scraped datasets)?
    \item[] Answer: \answerYes{} 
    \item[] Justification: Section 9 describes the controlled access policy, exclusion of high-evasion recipes, and restriction to synthetic data only.

\item {\bf Licenses for existing assets}
    \item[] Question: Are the creators or original owners of assets (e.g., code, data, models), used in the paper, properly credited and are the license and terms of use explicitly mentioned and properly respected?
    \item[] Answer: \answerYes{} 
    \item[] Justification: All referenced models and datasets are properly cited. No proprietary or scraped datasets were used.

\item {\bf New assets}
    \item[] Question: Are new assets introduced in the paper well documented and is the documentation provided alongside the assets?
    \item[] Answer: \answerYes{}
    \item[] Justification: The modulated dataset of 700 examples across seven strategies and five modulation levels is described in detail in Section 4.1. Controlled access is available to qualified researchers.

\item {\bf Crowdsourcing and research with human subjects}
    \item[] Question: For crowdsourcing experiments and research with human subjects, does the paper include the full text of instructions given to participants and screenshots, if applicable, as well as details about compensation (if any)? 
    \item[] Answer: \answerNA{}
    \item[] Justification: No human subjects were involved. All experiments were conducted using LLMs.

\item {\bf Institutional review board (IRB) approvals or equivalent for research with human subjects}
    \item[] Question: Does the paper describe potential risks incurred by study participants, whether such risks were disclosed to the subjects, and whether Institutional Review Board (IRB) approvals (or an equivalent approval/review based on the requirements of your country or institution) were obtained?
    \item[] Answer: \answerNA{}
    \item[] Justification: No human subjects research was conducted. All experiments used LLMs only.

\item {\bf Declaration of LLM usage}
    \item[] Question: Does the paper describe the usage of LLMs if it is an important, original, or non-standard component of the core methods in this research? Note that if the LLM is used only for writing, editing, or formatting purposes and does \emph{not} impact the core methodology, scientific rigor, or originality of the research, declaration is not required.
    \item[] Answer: \answerYes{}
    \item[] Justification: LLMs are a core and non-standard component of both the dataset construction and the experimental evaluation, described throughout Section 4.

\end{enumerate}

\end{document}